# Robot-Enabled Construction Assembly with Automated Sequence Planning based on ChatGPT: RoboGPT

Hengxu You[1], Yang Ye[2], Tianyu Zhou, Ph.D.[3], Qi Zhu,Ph.D.[4], Jing Du, Ph.D.[5]*
[1] Engineering School of Sustainable Infrastructure & Environment, University of Florida, Gainesville, FL 32611, USA
[2] Engineering School of Sustainable Infrastructure & Environment, University of Florida, Gainesville, FL 32611, USA
[3] Engineering School of Sustainable Infrastructure & Environment, University of Florida, Gainesville, FL 32611, USA
[4] National Institute of Standards and Technology, Boulder, CO 80305, USA
[5] Engineering School of Sustainable Infrastructure & Environment, University of Florida, Gainesville, FL 32611, USA

Corresponding author: Jing Du (e-mail: eric.du@essie.ufl.edu).

**ABSTRACT** Robot-based assembly in construction has emerged as a promising solution to address numerous challenges such as increasing costs, labor shortages, and the demand for safe and efficient construction processes. One of the main obstacles in realizing the full potential of these robotic systems is the need for effective and efficient sequence planning for construction tasks. Current approaches, including mathematical and heuristic techniques or machine learning methods, face limitations in their adaptability and scalability to dynamic construction environments. To expand the ability of the current robot system in sequential understanding, this paper introduces RoboGPT, a novel system that leverages the advanced reasoning capabilities of ChatGPT, a large language model, for automated sequence planning in robot-based assembly applied to construction tasks. The proposed system adapts ChatGPT for construction sequence planning and demonstrate its feasibility and effectiveness through experimental evaluation including Two case studies and 80 trials about real construction tasks. The results show that RoboGPT-driven robots can handle complex construction operations and adapt to changes on the fly. This paper contributes to the ongoing efforts to enhance the capabilities and performance of robot-based assembly systems in the construction industry, and it paves the way for further integration of large language model technologies in the field of construction robotics.

**INDEX TERMS** ChatGPT, Robot-assembly, Sequential Learning,

## I. INTRODUCTION

Robot-based construction assembly refers to the use of robotic systems for joining together various building components, materials, and systems to form a complete structure or a part of a structure [1]. It has emerged as a promising solution to address various challenges including increasing costs, labor shortages, project schedules, and the increasing demand for safe and efficient construction processes [2]. The use of robotic systems and the corresponding changes to the existing construction workflow are expected to significantly enhance productivity, reduce construction costs, and improve safety of construction projects [3]. Moreover, robot-based assembly systems can perform construction tasks that are repetitive, hazardous, or require high precision, thereby alleviating the burden on human workers [4].

Despite the potential benefits of robot-based assembly in construction, one of the main challenges faced by these systems is the need for effective and efficient sequence planning. A construction task often consists of a variety of interdependent steps that must be executed in a specific sequence order [5]. For example, installing a plumbing system requires a proper sequence of connecting pipes of different diameters and lengths, and using the appropriate couplings. Similarly, bricklaying requires placing the right bricks in the corresponding locations in the correct sequence. Many of these sequence planning tasks rely on spontaneous decisions, as construction tasks are often less predictable and difficult to plan out due to varying site conditions, resource availability, and evolving requirements [6]. As a result, construction workers often need to perform manual sequence planning on the fly, which involves determining the optimal order of construction steps and the corresponding logistics considerations. Manual sequence planning is a time-consuming and labor-intensive process, requiring a significant amount of experience to ensure the quality and accuracy.

Moreover, the complexity of construction task sequences can vary significantly depending on the specific construction project, further increasing the difficulty of the task [7]. Without an effective method for automated sequence planning, robot-based construction automation would not be scalable for meeting needs of real-world complex construction tasks.

In order to enable automation systems (including construction robotics) to handle more complex multi-step operational tasks, efforts have been made to explore heuristics-based methods or learning based methods. Early investigations included the use of mathematical and heuristic techniques in tackling the complex problem of sequence planning such as mixed-integer linear programming (MILP) (e.g., [8]). Recently, advances in machine learning are leveraged to support complex sequence planning with various constraints (e.g., [9]). These techniques aim to optimize operational sequences by considering factors such as precedence constraints, resource availability, and task interdependencies. By integrating these approaches with robotic systems, researchers expect to develop more efficient and adaptable solutions that can manage the inherent complexities and uncertainties of construction operations.

However, these methods have certain limitations that hinder their effectiveness in addressing the dynamic nature of construction projects. On the one hand, mathematical and heuristic techniques often involve the development of tailored algorithms (by human experts) that leverage domain-specific knowledge and rules [10]. While these methods can effectively navigate the complex solution space for complex and variable construction tasks, they may impose a significant computational overhead due to the need for continuous adaptation and refinement of the heuristics as the construction process evolves. On the other hand, although machine learning methods, such as genetic algorithms and neural networks, can adapt to dynamic scenarios much easier compared to mathematical and heuristic techniques, they require a significant amount of training data to achieve accurate results [11]. In construction operations, where site conditions and project requirements can change frequently, acquiring sufficient training data for every possible scenario is challenging, limiting the adaptability of these methods to dynamic environments.

In this paper, we introduce the novel system called RoboGPT that utilizes ChatGPT for automated sequence planning in robot-based assembly applied to construction tasks. ChatGPT, as an advanced large language models (LLMs), has demonstrated remarkable capabilities in understanding and generating human-like text, which rely on a reasoning ability for understanding the inherent structures of a sequence [12]. This paper hypothesizes that the reasoning ability of ChatGPT can be leveraged for developing an efficient and adaptable sequence planning algorithm. By integrating ChatGPT into the construction process, we aim to minimize the reliance on manual intervention, reduce planning time, and increase the overall efficiency of robot-based assembly systems in the construction industry.

Specifically, in this paper we will show how we adapted ChatGPT for the purpose of automated sequence planning in robot-based assembly for construction applications and demonstrate the feasibility and effectiveness of the proposed approach through an experimental evaluation, including comparing the ability of ChatGPT-driven robots in handling complex construction operations and adapt to changes on the fly. By accomplishing these goals, this paper will contribute to the ongoing efforts to enhance the capabilities and performance of robot-based assembly systems in the construction industry and pave the way for further integration of LLMs technologies in the field of construction robotics.

## II. LITERATURE REVIEW

### A. Construction Robotics for Assembly Tasks

In recent years, the adoption of robotics in the construction industry has grown, aiming to improve efficiency, reduce labor costs, and enhance safety on construction sites. This literature review explores various robotic systems and approaches that have been developed for construction assembly tasks, highlighting their advantages and challenges. Early research in construction robotics focused on developing specialized robotic systems for specific tasks. One example is the masonry robot SAM (Semi-Automated Mason) developed by Construction Robotics [13], which automates the bricklaying process, leading to reduced labor costs and increased productivity. Another example is the Ty Bot by Advanced Construction Robotics, a rebar tying robot that streamlines the reinforcement process in concrete construction [14]. However, these specialized robotic systems often lack the flexibility to adapt to the dynamic and complex nature of construction environments. As a result, researchers have explored the use of more versatile robotic systems, such as modular robots and robotic arms, which can be reconfigured and programmed to perform various construction tasks. One notable example is a modular robotic system capable of autonomously assembling truss structures [15]. Similarly, Apolinarska, Pacher, Li, Cote, Pastrana, Gramazio and Kohler [16] demonstrated the use of an industrial robotic arm in assembling complex timber structures, highlighting the potential for large-scale applications of robotic arms in construction. The emergence of digital fabrication techniques, such as 3D printing, has also influenced the development of construction robotics. A well-known example is the MX3D project, which utilized a robotic arm to 3D print a steel pedestrian bridge in Amsterdam [17]. Another study by Oke, Atofarati and Bello [18] presented the Digital Construction Platform (DCP), a mobile robotic system capable of 3D printing building components on-site, offering a flexible and scalable approach to automated construction. Integration of robotic systems with Building Information Modeling (BIM) has been another area of interest for researchers. BIM, as a digital representation of a building's physical and functional characteristics, provides a wealth of information that can be utilized by robotic systems to plan and execute construction tasks. For instance, Gao, Meng, Shu and Liu [19] proposed a BIM-based robotic assembly system for prefabricated building components, demonstrating the potential for enhanced efficiency and accuracy in construction processes. Collaborative robotics is another important area in construction assembly tasks, where multiple robots work together to achieve a common goal. Carey, Bardunias, Nagpal and Werfel [20] demonstrated a swarm of construction robots inspired by termite behavior that could collaboratively build structures without centralized control. Similarly, Ding, Dwivedi and Kovacevic [21] showcased a multi-robotic system for wire arc additive manufacturing, highlighting the advantages of distributed robotic systems in construction.

Despite the advances in construction robotics, several challenges remain, including the need for robust perception and decision-making capabilities. To address these challenges, researchers have explored the integration of advanced computer vision and artificial intelligence (AI) techniques. For example, Zhang, Shen and Li [22] developed a computer vision-based method for autonomous rebar picking and placing using a robotic arm, while Osa and Aizawa [23] demonstrated the use of deep learning algorithms for automating excavation tasks. Another primary challenge is the adaptability and versatility of robotic systems in varying construction environments. As stated by Ardiny, Witwicki and Mondada [24], construction sites are dynamic and often unpredictable, making it difficult for robotic systems to perform tasks efficiently without constant human intervention. Another limitation is the high cost of developing and implementing advanced robotic systems, which may not be feasible for smaller construction firms. Furthermore, the integration of robotic systems requires extensive training for construction workers, which can be time-consuming and costly [25]. Moreover, robotic systems often struggle with tasks that must be conducted with a proper sequence order of a series of steps, or assembly sequence planning (ASP) [26], which is the main focus of our investigation and will be discussed in depth in the next section.

*B. Sequence Planning for Multi-Step Operations*

Sequence planning for multi-step operations is largely addressed by the assembly sequence planning (ASP) literature, which is a critical aspect of manufacturing and construction that aims to identify the most efficient and cost-effective sequence of operations to assemble a product while considering various factors, such as resources, constraints, and goals [26]. Classical approaches to ASP include graph-based methods, such as the AND/OR graph [27] and the liaison graph [28], which represent assembly operations as directed graphs with nodes representing parts and edges representing assembly operations. Matrix-based methods, such as the design structure matrix (DSM) [29] and the assembly incidence matrix (AIM) [30], use matrices to represent relationships between parts and assembly operations. Expert systems, like the blackboard system [31] and the CLIPS-based approach [32], employ human expert knowledge in the form of rules to generate assembly sequences. Additionally, mathematical methods like integer programming (IP) [33], mixed-integer linear programming (MILP) [8], and constraint programming (CP) [34] have been used to model and solve ASP problems by formulating them as mathematical models with variables representing assembly operations and constraints representing precedence relationships and resource limitations. Heuristic methods, such as greedy algorithms [35], local search methods like simulated annealing [36], tabu search [37], and variable neighborhood search [38], constructive heuristics like the minimum degree heuristic [39], and decomposition methods [40] have been employed to find good solutions efficiently by using simple rules and shortcuts to navigate the complex solution space of ASP problems. These heuristic methods are typically faster and less computationally demanding compared to optimization techniques, making them suitable for large-scale assembly problems. In recent years, modern approaches leveraging computational power and artificial intelligence have emerged, such as genetic algorithms [41], ant colony optimization [42], particle swarm optimization [43], and artificial neural networks [9]. These methods are used to explore a wide search space and generate optimal or near-optimal solutions.

However, these methods have certain limitations that hinder their effectiveness in addressing the dynamic nature of construction projects. Most ASP methods assume deterministic input data and fail to handle uncertainties related to structure dimensions, assembly resources, and process variability. In real-world production environments (e.g., manufacturing and construction), various sources of uncertainty may arise, such as geometric and material property variations, tool capabilities, resource availability, and human operator performance [44]. As such, to facilitate construction tasks sequence planning, mathematical and heuristic techniques will need to rely on tailored algorithms by human experts that leverage domain-specific knowledge and rules. While these methods can effectively navigate the complex solution space for complex and variable construction tasks, they may impose a significant computational overhead due to the need for continuous adaptation and refinement of the heuristics as the construction process evolves [45]. This limitation may lead to suboptimal solutions in highly dynamic and uncertain environments, where rapid adjustments to changing circumstances are essential [46]. On the other hand, although machine learning methods, such as genetic algorithms and neural networks, can adapt to dynamic scenarios much easier compared to mathematical and heuristic techniques, they require a significant amount of training data to achieve accurate results [11]. In construction operations, where site conditions and project requirements can change frequently, acquiring sufficient training data for every possible scenario is challenging, limiting the adaptability of these methods to dynamic environments [47]. Moreover, the performance of machine learning methods may be negatively affected by the presence of noisy or incomplete data [48], which is often the case in construction projects.

*C. LLMs for Sequence Planning*

Large language models (LLMs), such as OpenAI's GPT series [49], BERT [50], and T5 [51], have demonstrated impressive performance in natural language understanding, generation, and reasoning tasks. These models, based on the Transformer architecture [52], utilize deep learning techniques and massive datasets to learn contextual representations of text, allowing them to generate coherent and contextually relevant responses. The application of large language models in sequence planning tasks, such as ASP, remains relatively unexplored. However, recent studies have started investigating the potential of these models for various sequence planning tasks. For example, LLMs have been used to generate textual descriptions of process plans based on given input constraints [53]. These models can potentially be employed to generate high-level assembly plans or provide guidance to human operators during the assembly process. In domains like project management and logistics, large language models have been used to generate natural language descriptions of optimal schedules or resource allocations [54]. Large language models have demonstrated potential in generating task plans for robotic systems based on natural language instructions [55]. These models can be adapted to generate action sequences for robotic assembly operations or other complex robotic tasks, providing a more intuitive human-robot interaction experience. This application can be extended to sequence planning tasks, such as generating assembly schedules or allocating resources for assembly operations. In a recent study by Prieto, Mengiste and García de Soto [54], ChatGPT-3.5 was used to generate a construction schedule for a simple construction project. The output from ChatGPT was evaluated by a pool of participants who provided feedback on their overall interaction experience and the quality of the output. The results showed that ChatGPT could generate a coherent schedule that followed a logical approach to fulfill the requirements of the scope indicated. The participants had an overall positive interaction experience and indicated the potential of such a tool in automating many preliminary and time-consuming tasks. Vemprala, Bonatti, Bucker and Kapoor [56] conducted an experimental study exploring the use of OpenAI's ChatGPT for robotics applications. The study proposed a strategy that combines design principles for prompt engineering and the creation of a high-level function library, enabling ChatGPT to adapt to different robotics tasks, simulators, and form factors. The evaluations focused on the effectiveness of various prompt engineering techniques and dialogue strategies in executing diverse robotics tasks.

As for the reasons why ChatGPT can be used for sequence planning are still largely unknown. Earlier evidence shows that ChatGPT has inherent capabilities as a large-scale language model, as it has been trained on massive amounts of text data, enabling it to understand and generate human-like text [12]. This capability may allow it to comprehend the requirements of a sequence planning problem and generate potential solutions in a human-readable format. In addition, it seems that ChatGPT can capture context and reason about the relationships between different elements in a sequence planning problem. It can consider constraints, dependencies, and objectives to generate effective solutions [57]. ChatGPT's extensive training data allows it to possess a broad range of knowledge across various domains. This knowledge could be potentially leveraged to address sequence planning tasks in multiple industries, such as manufacturing, construction, logistics, and robotics [58]. The exploration of this study in applying ChatGPT in complex sequence planning for robotics is expected to add more empirical evidence and new methods for automation.

## III. SYSTEM DESIGN

*A. Architecture*

**Fig.1** presents the comprehensive system architecture of RoboGPT, which is composed of four primary components: Robot Control System, Scene Semantic System, Objects Matching System, and User Command Decoder System. ChatGPT, an advanced natural language processing model, functions as the central intelligence within the system. Upon receiving task descriptions and specific requirements from users, ChatGPT meticulously generates sequential solution commands in a step-by-step manner, adhering to the precise requirements of the task. The generated response text is subsequently decoded by the User Command Decoder System and transmitted to a Unity-based virtual environment in the form of virtual objects. The Scene Semantic System is responsible for detecting real-world objects, which are then sent to the Unity environment to be meticulously aligned and matched with their virtual counterparts. Once the alignment is complete, the objects, in conjunction with the corresponding actions derived from the commands, are relayed to the Robot Control System to facilitate real-world object manipulation.

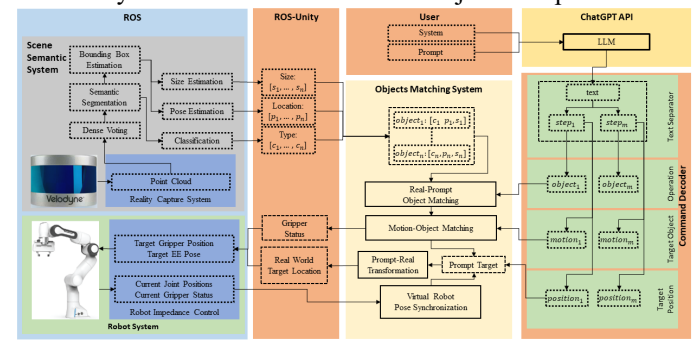

**FIGURE 1.** System Structure

*B. Robot Control System*

The robot system tested in this study was a Franka Emika Panda robot arm, a lightweight, compact, and versatile robot designed for human-robot collaboration, which is widely used in manufacturing, research, and education, and is known for its ease of use, flexibility, and reliability. The Panda has seven degrees of freedom corresponding to its seven joints. Each joint is equipped with a force/torque sensor and a joint angle sensor to accurately measure the robot arm states, allowing it to move in various directions and perform intricate tasks with high precision. A Parallel gripper is attached as the end effector on the seventh joint that could be used to interact with object such picking and dropping.

To smoothly control the end effector and generate a stable moving trajectory, the impedance controller in cartesian coordinates is applied as shown in Figure 2. The impedance of the end-effector can be adjusted based on the force or torque applied by the environment, allowing the robot to adapt to varying conditions. Specifically, the controller imposes a spring-mass-damper behavior on the mechanism by maintaining a dynamic relationship between force and position, velocity and acceleration:

$$\boldsymbol{F} = C\boldsymbol{v}_{ee} + K\Delta \boldsymbol{x}_{ee}^{rob} + \boldsymbol{ld}, \quad (1)$$

where $\boldsymbol{F}, \boldsymbol{v}_{ee}, \Delta \boldsymbol{x}_{ee}^{rob}, \boldsymbol{ld} \in \mathbb{R}^3$ are the implemented force on the end effector, velocity of the end effector, position of the end effector in the robot coordinate system and payload, respectively. Given the end effector's current position $\boldsymbol{x}_{ee\_curr}^{rob}$ and the desired position $\boldsymbol{x}_{ee\_desire}^{rob}$, $\Delta \boldsymbol{x}_{ee}^{rob}$ is calculated as:

$$\Delta \boldsymbol{x}_{ee}^{rob} = \boldsymbol{x}_{ee_{desire}}^{rob} - \boldsymbol{x}_{ee_{curr}}^{rob}. \quad (2)$$

$\boldsymbol{x}_{ee\_desire}^{rob}$ is the real-world target location derived from the Real-Virtual Objects Matching System that is going to be discussed in the following section. In order to control the virtual robot arm in Unity to interact with the virtual objects, the real-time joint position $\boldsymbol{q}^{rob} \in \mathbb{R}^7$ is sent to Unity through the ROS-Unity bridge (RUB) to synchronize the virtual arm. Each element in $\boldsymbol{q}^{rob}$ is the rotation angle for the corresponding joint. The gripper's status of the real robot arm is also sent through RUB to instruct the virtual robot's behavior and the interaction between the virtual gripper and objects will be sent back to ROS to control the real gripper's action.

*C. Semantic Segmentation System*

The Scene Semantic System collects the visual information from surrounding environment and detects the real target objects for downstream alignment. A Velodyne-16 LiDAR (VL16) is used to capture point cloud data and save it on the ROS platform. The LiDAR sensor coordinate system is calibrated with the Panda coordinate system to ensure that the positions of detected objects.

The VL16 is selected as the scanning sensor because of its high scanning speed and stable scanning results. Since the VL16 only have 16 scanning rings in the vertical direction which is too sparse to capture the detail spatial information, an augmentation scanning strategy [ref: paper 0] is applied to register the scanning results from multiple viewpoints and generate a dense scanning result. To eliminate the influence of the error caused by multiple frames registration, we apply the density-voting clustering method to shift the drifting points to its closest density center so that all the returning points would be close to the object surfaces and the shape of objects could be perfectly captured.

The virtual scene data, including joint states, point cloud, and virtual objects with physical properties, is then sent to engine the Unity game for interface reconstruction. In order to subscribe data from ROS via the network, ROS-Unity bridge and ROS# are used to build a WebSocket, which allows two-way communication between ROS and Unity data transfer. We also used ROS# to build some nodes in Unity to publish and subscribe topics from ROS. Baxter's state data (URDF, joint, and gripper state) is used to build a virtual Baxter that replicates the same states of the real Baxter. The same prefab library as mentioned in the scene recognition system is used to provide virtual object information with physical properties that can be used to rebuild stationary objects in the game engine. We also use Unity physical engine to assign the point cloud and virtual object with physical proprieties and rebuild a virtual working scene based on the data from ROS.

The augmented and clustered point cloud $PC^{cam} \in \mathbb{R}^{N*3}$ is then fed to PointNet++ [ref: PN++] which we take as our segmentation model as shown in Figure 2. N denotes the number of points according to the input size of the model. The PointNet++ is a well-trained deep learning model on various of point cloud dataset and could handle both the object detection and semantic segmentation tasks. In this application, we only focus on the segmentation branch of PointNet++ to get the object labels of each point. The segmented points are clustered as point sets $[PC_0^{cam}, \ldots, PC_n^{cam}]$ and the corresponding predicted labels $[c_0, \ldots, c_n]$. The point sets are then used to estimate the oriented bounding boxes that closely wrap all the points as $[Box_0^{cam}, \ldots, Box_n^{cam}]$. The bounding boxes are parameterized as $Box_i^{cam} := [\boldsymbol{s}_i^T, \boldsymbol{p}_i^T]^T$, where $\boldsymbol{s}_i, \boldsymbol{p}_i \in \mathbb{R}^3$ are the size (width, length, height) and location ($x_i, y_i, z_i$). The labels, sizes and locations of segmented point sets are then sent to Unity through RUB as classification results, size estimation results and pose estimation results.

*D. Command Decoder*

The command decoder works as the translator to transfer the response from ChatGPT in natural language into machine-understandable programming command so that the robot arm could execute the actual sequential actions inferred by ChatGPT. We used the ChatGPT-4 model and coded with python and C# to build the API to communicate between Unity and the online model. The API is based on HTTP request. The user sends a text prompt to the API, and it will return a response in the form of a text message. The API also supports various customization options to regularize the response by typing the specific requirements in the "system" section.

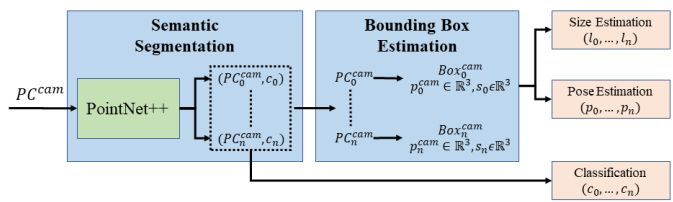

**FIGURE 2.** Scene Semantic System.

For most construction assembly task, the sequential actions could be simplified as moving an object to a certain location. For example, moving the pipe to position A or put the brick to position B. Therefore, the operation command for the sequential actions could by represented by action, object and target position. In order to make the reply from ChatGPT more explainable, we set the "system" with three principles:

1) *The ChatGPT will generate the reply step by step in an execution order.*

2) *For each step, there is only one motion and one object to be moved or operated. There is only one target location.*

3) *The related words about action, object and target position, must be quoted by brackets.*

Given the regularization principles, the reply from ChatGPT could be simplified as:

*Step 1. [Action 1] [object 1] to [position 1].*
*Step 2. [Action 2] [object 2] to [position 2].*
*…*
*Step n. [Action n] [object n] to [position n].*

Therefore, the regularized reply from ChatGPT could be firstly split into single steps. Then the single steps could be used to extract action, object and target position as shown in **Fig.3** The brackets are used to crop the action or object names as strings. The detected string will then be checked to see if it shows up in the pre-defined action or object dictionary. If the dictionaries contain the string, the corresponding action or object will be sent to the Real-Virtual Objects Matching System. The dictionary contains the name of common actions and objects in construction sites.

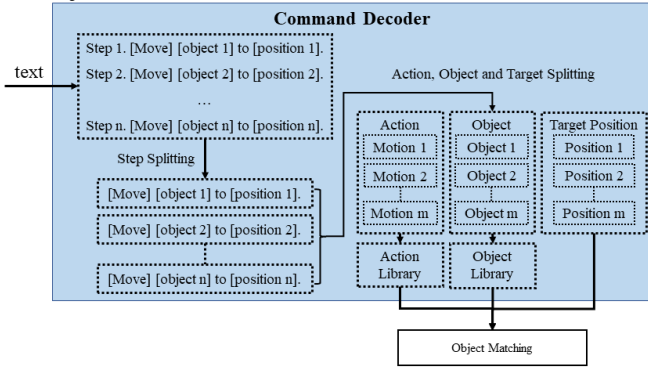

FIGURE 3. Command Decoder Structure

*E. Object Matching System*

The detected action, object and position will then be sent to the matching system to be paired with the detected objects from real world and be transferred to robot arm control codes. Specifically, the detected objected, noted as $obj^{prompt}$, will be firstly matched with the label of segmented objects from Semantic Segmentation System, noted as $obj^{seg}$. Note that the labels of the segmentation system are strings that included in the object dictionary. Given a matched pair $(obj_i^{prompt}, obj_j^{seg})$, $obj_i^{prompt}$ is then assigned the parameters of $obj_j^{seg}$ including $l_j^{seg}$ for size and $p_j^{seg}$ for position. Then $obj_i^{prompt}$ has four major properties as:

$$obj_i^{prompt} := [l_j^{seg}, p_j^{seg}, action_j^{prompt}, position_i^{prompt}], \quad (3)$$

where $action_j^{prompt}$ is the matched action from the dictionary and $position_i^{prompt}$ is the target location. Thus, the desired operation on $obj_i^{prompt}$ are parameterized as its current position, the action and its target position, which could be understood by the robot arm. Then $obj_i^{prompt}$ could be sent to the Panda for a single-step operation in a sequence.

**IV. TEST CASES**

In this section, we report two cases to showcase how the proposed RoboGPT system accomplished generic assembly tasks, including material stacking and Hanoi tower puzzle. The detailed results of the test cases are as follows.

*A. Material Stacking*

The first case, object stacking, related to placing materials based on their sizes. A real-world example can be stacking construction materials on the site based on their dimensions and shapes for a stable structure. We used four disks of different sizes as an example. The proposed system was requested to stack the disks as stable as possible. The prompt we gave to the ChatGPT was: *I have five cubes with names [A] to [E]. The cubes' lengths are in a descending order from [A] to [E]. So, I want to teach a robot arm to use the cubes to create a tower with the most stable design. Could you tell me which cube to operate step by step?*

ChatGPT analyzed the size and weight of the disks and generated an assembly sequence, starting with the largest disk at the bottom and placing the others in decreasing order of size, as follows:

1. *Move cube [A] to the base location where the tower will be built.*
2. *Move cube [B] from its original location to the top of cube [A].*
3. *Move cube [C] from its original location to the top of cube [B].*
4. *Move cube [D] from its original location to the top of cube [C].*
5. *Move cube [E] from its original location to the top of cube [D].*

The UI in Unity was shown in **Fig.4**. Clicking the button on the top right will send the prompt from Unity to Python ChatGPT-4 API as shown in **Fig.5**. Then, the deployed ChatGPT-4 API sent the generated assembly sequence to a robot arm (Franka Emika Panda) for locomotion controls.

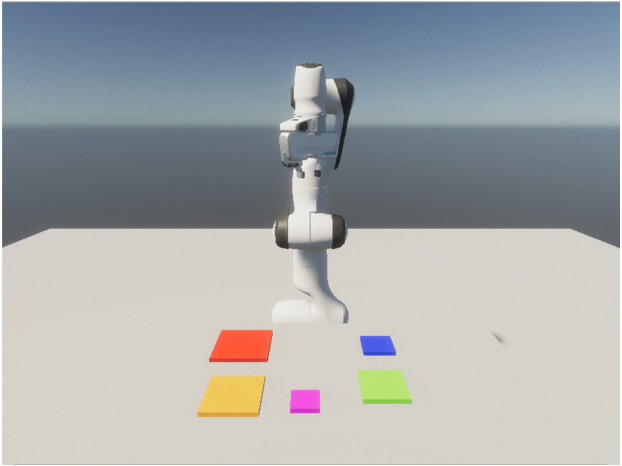

FIGURE 4. ChatGPT-4 user interface in Unity for material stacking.

FIGURE 5. ChatGPT-4 response for robot control in material stacking.

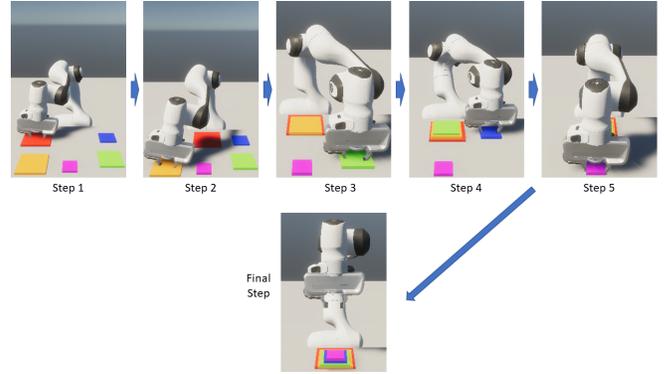

FIGURE 6. Simulated robot arm building the 5-cubes tower.

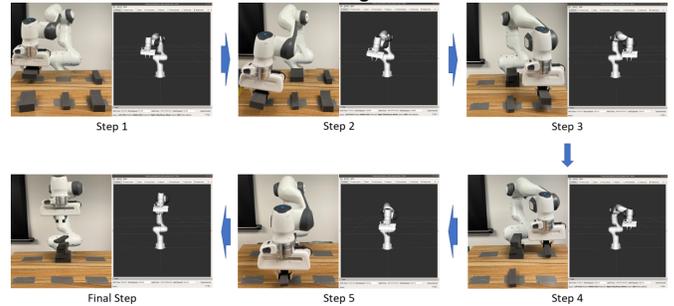

FIGURE 7. Real robot arm stacking 5 soft blocks.

The computer vision module, utilizing a combination of object detection and recognition algorithms, was used to identify the five disks and their positions in the workspace. We simply used five obvious labels to help the robot arm to locate the objects since we didn't focus on building object detection algorithms. With this information, the system calculated the necessary movements for the robot arm to execute the stacking task based on the described sequence order. By precisely controlling the position, orientation, and gripping force of the robot arm, the system successfully stacked the five cubes in a stable arrangement, demonstrating the effectiveness of ChatGPT-4 in guiding robotic systems to accomplish complex assembly tasks. **Fig.6** showed the simulated robot arm placing the objects in Unity. Figure 7 showed the Franka Emika Panda robot arm placing blocks to form a tower in real world. For each step, the sub-figure on the left showed the real robot arm's action and the sub-figure on the right showed the pose of robot arm in ROS.

### B. Hanoi Tower Puzzle

The second case we tested was the classic Tower of Hanoi puzzle with four disks of different sizes. ChatGPT was used to generate an optimal solution, which involved moving the disks among three pegs while adhering to the puzzle's rules: only one disk can be moved at a time, and a disk cannot be placed on top of a smaller disk. The prompt we provided was: *I have a tower of Hanoi with five disks [A], [B], [C], [D], [E], from smallest to biggest. Describe the sequence of completing the puzzle and control the robot arm to finish it.*

ChatGPT formulated a step-by-step sequence to solve the puzzle, guiding the robot to move the disks between the pegs in the correct order. The robot successfully completed the task, demonstrating the ability of ChatGPT to solve complex problems with specific constraints. In this case, the ChatGPT-4 API was deployed to solve a Hanoi tower puzzle using a robot arm (Franka Emika Panda). **Fig.8** showed the same UI in Unity and the setup for Hanoi Tower Puzzle problem. **Fig.9** showed the prompt sent from Unity to Python and the sequence replied by ChatGPT. By accurately controlling the position, orientation, and gripping force of the robot arm, the system was able to successfully stack the four disks in a stable arrangement.

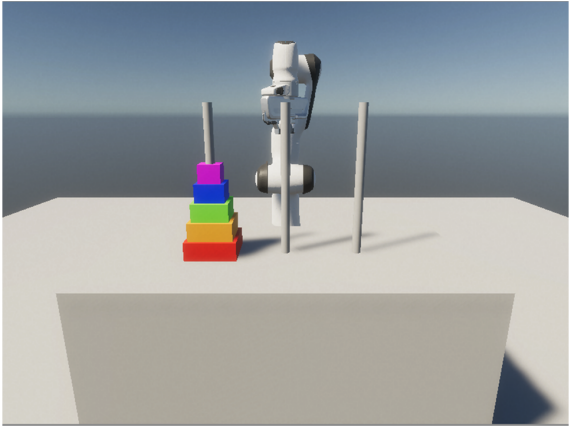

**FIGURE 8.** ChatGPT-4 user interface in Unity for Hanoi Tower Puzzle.

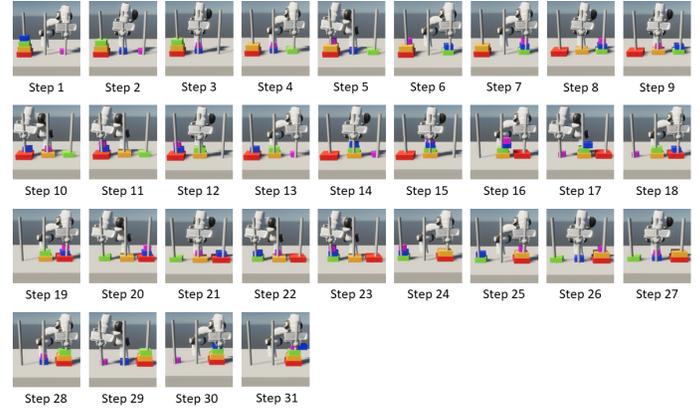

**FIGURE 10.** Simulated robot arm solving the Hanoi Tower Puzzle.

**FIGURE 9.** ChatGPT-4 response for robot control in Hanoi Tower Puzzle.

The generated assembly sequence was sent to the robot arm for locomotion controls. Similarly, a computer vision module, consisting of object detection and recognition algorithms, was employed to identify the fives disks and the three towers along with their positions within the workspace. To simplify the system design, we used three square labels to represent the towers and only required the real robot arm to recognize the location of the towers. Upon obtaining the positions of the disks, the system calculated the necessary robot arm movements to execute the stacking task according to the sequence order provided by ChatGPT-4. **Fig.10** showed the step-by-step motion of the simulated robot arm in Unity to solve the 5-disks Hanoi Tower Puzzle. The response from ChatGPT-4 was the optimal answer as a 31-steps sequence and the robot arm could precisely follow the instruction from ChatGPT and completed the problem. This test further proved that the proposed RoboGPT system could effectively solve the sequential learning problem and interact with the real objects to solve the problem in real world.

## V. COMPARISON STUDY

In order to demonstrate the advantages of the proposed RoboGPT system in intricate multi-stage robotic operations and investigate the capacity of ChatGPT to address real-world construction challenges, we conducted a comprehensive evaluation of the RoboGPT system in the context of the pipeline installation under various conditions. This comparative study aimed to assess the system's performance, as well as to elucidate its potential and limitations under different conditions, such as the variability of the raw materials and the task requirements.

We opted not to incorporate the material stacking and Hanoi tower puzzle scenarios in this comparison investigation for two primary reasons. Firstly, the material stacking task is relatively elementary, as it predominantly necessitates rudimentary knowledge of object stacking based on size. Secondly, the central challenge of the Hanoi tower puzzle resides in solving the puzzle within a constrained timeframe, which has been well addressed by other algorithms, and does not align with the objectives of our study.

Conversely, the pipeline installation scenario presented a more open-ended challenge, requiring the system to determine the spatial dimensions, evaluate resource availability, and devise an appropriate method for connecting the pipes. It is crucial to note that this task does not entail a singular solution; rather, multiple viable solutions can achieve the desired outcome. Consequently, the pipeline installation task, which demands a thorough assessment of dimensions, resource estimation, and sequencing while considering both spatial and resource constraints, is better suited for our comparative analysis. We applied two different tasks with two different conditions to evaluate the performance of the purposed system. Since the pipe installation tasks in real world often required large spatial space which was hard to manipulate with a research-based robot arm, we built the simulation environment in Unity to test the results. Given the knowledge from pipe installation process in real construction site, we designed the *Avoid Obstacles* and *Pass Points* tasks for further testing.

The *Avoid Obstacles* task was still to design the pipeline to connect two points, but the pipes cannot pass certain points. This task was designed to simulate the case that the pipes have avoid some pre-built structures or safety areas. The testing environment was designed as a 10*10*10 room with the start point location to be $P^1_{start} = (5,5,0)$ at the floor and the end point to be $P^1_{end} = (5,5,10)$ at the roof. The two obstacle points, named $A_{obs}$ and $B_{obs}$, were located at $(5,5,5)$ and $(5,7,5)$, respectively. **Fig.11** showed the setup environment of *Avoid Obstacles*. The green cube denoted the start point and the red cube denoted the end point. The two small black cubes denoted the obstacles to be avoided.

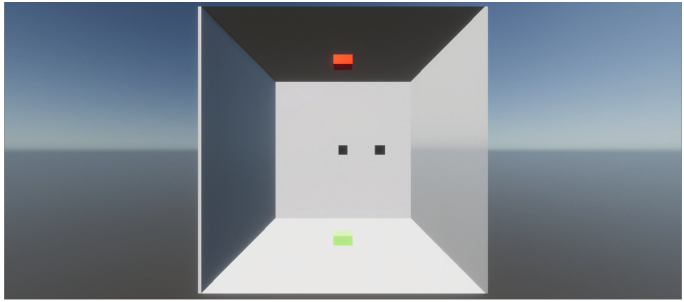

**FIGURE 11.** The setup environment of Avoid Obstacles task.

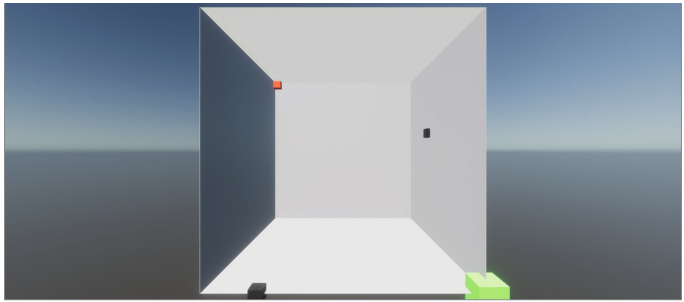

**FIGURE 12.** The setup environment of Pass Points task.

The *Pass Points* task is to find a solution to design the pipeline between two given positions and pass certain points. This situation was designed to simulate the case that the pipes must connect some devices such as the air conditioners or the pipes must pass through some holders on the wall as supports. Similarly, the testing environment was also in a 10*10*10 room with the start point location to be $P^2_{start} = (0,0,0)$ and the end point to be $P^2_{end} = (10,10,10)$. The two mandatory points, named $A_{man}$ and $B_{man}$, were located at $(0,0,8)$ and $(6,6,0)$, respectively. **Fig.12** showed the setup environment of *Pass Points*. Similarly, the green cube denoted the start point and the red cube denoted the end point. The two small black cubes denoted the mandatory points to be connected.

### A. Avoid Obstacles Task

In the *Avoid Obstacles* task, we made two different conditions: constant condition and variable condition. To be Specific, the constant condition referred to the situation that the pipes to be used to build the pipeline were with the same size. In our case, we set the length of pipe to be 2. Note that the diameter of the pipe was ignored. On the contrary, the variable condition referred to the case that the pipes' sizes were not fixed. To be specific, we use three types of pipes with the length of 2, 3, and 4 respectively. The system can choose any of the pipes to build the pipeline.

The prompt we used for the constant condition was listed below: *Can you help me with pipe connection? We have several 2ft length straight pipes (pipe 2ft), 3ft length straight pipes (pipe 3ft), 4ft length straight pipes (pipe 4ft). The start position is (5ft, 5ft, 0ft) direction is the positive Z axis, the end position (5, 5, 10) direction is the negative Z axis. We assume that each straight pipe can be connect to each other directly. You can just tell me the position of each pipe, such as 'pipe 2ft #1 (5, 5, 2) z axis, pipe 2ft #2 (5, 5, 4) z axis, pipe 2ft #3 (5, 7, 4) y axis'. To be noted, each pipe must maintain parallelism to the X, Y, and Z axes. There are two obstacles at point (5, 5, 5) and point (5, 7, 5), the pipe cannot pass through this point from neither X, Y nor Z axes.*

The prompt for the variable condition is: *Can you help me with pipe connection? We have several 2ft length straight pipes (pipe 2ft), 3ft length straight pipes (pipe 3ft), 4ft length straight pipes (pipe 4ft). The start position is (0ft, 0ft, 0ft) direction is the positive Z axis, the pipe connection must pass the first mandatory point (0, 0, 8), then pass the second mandatory point (6, 6, 0), finally to the end position (10, 10, 10) direction is the negative Z axis. We assume that each straight pipe can be connect to each other directly. You can just tell me the position of each pipe, such as 'pipe 2ft #1 (0, 0, 2) z axis, pipe 4ft #1 (0, 0, 6) z axis, pipe 3ft #1 (0, 3, 6) y axis'. To be noted, each pipe must maintain parallelism to the X, Y, and Z axes. The pipe must pass each mandatory point (0, 0, 8) and (6, 6, 0).*

For each condition, we used the same prompt to generate 20 trials. **Table.I** listed the counting results of successful and failed trials. The sub-optimal trials referred to the case that the RoboGPT system could give the correct connection design but with unnecessary pipes and detour.

TABLE 1
THE COUNTING RESULTS OF EXPERIMENTS ON AVOID OBSTACLES TASK WITH TWO CONDITIONS.

|  | Success | | Fail | Total | Optimal/Total | Success/Total | Failed/Total |
|---|---|---|---|---|---|---|---|
|  | Optimal | Sub-optimal |  |  |  |  |  |
| Constant | 12 | 8 | 0 | 20 | 0.6 | 1 | 0 |
| Variable | 3 | 5 | 12 | 20 | 0.15 | 0.25 | 0.8 |

The results showed the significant difference between the successful rates of the two conditions as 100% for constant condition and 25% for variable condition. Theoretically, the two conditions were corresponding to two difficulty levels on solving the problem. For the first condition, the pipe's length is almost the unit length compared with the room's scale. There is no need to consider the arrangement of pipes to fit certain length of the total pipeline. In other words, the final solution could use any number of pipes and the only requirement was to avoid $A_{obs}$, $B_{obs}$ and finally reach $P_{end}^1$. However, for the second condition, the length of pipe varies from 2 to 4. So, the solution had to not only satisfy the requirement to pass the mandatory points and reach target but also find the proper combination of pipes with different size to fit the length of the pipeline. There was an extra constraint which restricted the solution space, added logical difficulty and made the problem harder to solve. In other word, the resource pipes that could be used to build the pipeline are restricted. **Fig.13** showed the assembling process of a successful trial.

**Fig.14** showed a typical sub-optimal solution that given sufficient pipes without any constraints, ChatGPT could give redundant design with unnecessary cost. The proposed pipeline by ChatGPT was making unnecessary detour to avoid the obstacles. **Fig.16** illustrated the shortage of ChatGPT in spatial understanding. The layout on the left showed the failure in condition 2 that the pipe only reach the height of end point but couldn't find the location on x-z plane. The failed layout on the right showed that the start and end point of a pipe was not understood so the following pipe were connected from the middle of the previous pipe as shown in the red circle. The results proved that the ChatGPT couldn't always precisely understand the spatial information from pure text input.

**FIGURE 13.** The assembling process of a successful trial for Avoid Obstacles task.

**FIGURE 14.** The layout of sub-optimal trial in constant condition.

**FIGURE 15.** The layout of successful trial in variable condition

**FIGURE 16.** The layout of failed trial in variable condition.

### B. Pass Points Task

In the *Pass Points* task, we used the same two conditions as in the previous task. The prompt we used for the constant condition was listed below: *Can you help me with pipe connection? We have several 2ft length straight pipes (pipe 2ft). The start position is (0ft, 0ft, 0ft) direction is the positive Z axis, the pipe connection must pass the first mandatory point (0, 0, 8), then pass the second mandatory point (6, 6, 0), finally to the end position (10, 10, 10) direction is the negative Z axis. We assume that each straight pipe can be connect to each other directly. You can just tell me the position of each pipe, such as 'pipe 2ft #1 (0, 0, 2) z axis, pipe 2ft #2 (0, 0, 4) z axis, pipe 2ft #3 (0, 2, 4) y axis'. To be noted, each pipe must maintain parallelism to the X, Y, and Z axes. The pipe must pass each mandatory point (0, 0, 8) and (6, 6, 0).*

The prompt for the variable condition is: *Can you help me with pipe connection? We have several 2ft length straight pipes (pipe 2ft), 3ft length straight pipes (pipe 3ft), 4ft length straight pipes (pipe 4ft). The start position is (0ft, 0ft, 0ft) direction is the positive Z axis, the pipe connection must pass the first mandatory point (0, 0, 8), then pass the second mandatory point (6, 6, 0), finally to the end position (10, 10, 10) direction is the negative Z axis. We assume that each straight pipe can be connect to each other directly. You can just tell me the position of each pipe, such as 'pipe 2ft #1 (0, 0, 2) z axis, pipe 4ft #1 (0, 0, 6) z axis, pipe 3ft #1 (0, 3, 6) y axis'. To be noted, each pipe must maintain parallelism to the X, Y, and Z axes. The pipe must pass each mandatory point (0, 0, 8) and (6, 6, 0).*

Similarly, we used the same prompt to generate 20 trials with new chat channel. **Table. II** listed the counting results of successful and failed trials. To intuitively show the results from the two conditions, we picked a success trial and a failed trail from each condition and give the visualization in **Fig.17-20**.

**Fig.17-18** showed the successful and failed trials of constant conditions. The layout in Figure 18 further proved the shortage of ChatGPT in spatial understanding. There were two gaps along the pipeline, indicating that ChatGPT might wrongly overlap the two points only based on their 2D coordinates. The two endpoints in the red circle had the same *x* and *z* coordinates but different *y* coordinates. The ones in the yellow circle had the same *x* and *y* coordinates but different *y* coordinates. In other words, ff the coordinates of two points were the same along one or two axes, they would be wrongly aligned and treated as the same points. In this case, it is reasonable to deduce that ChatGPT relied more on pure separated numerical analysis to solve the real-world problem. The x-, y- and z-coordinates of the two points might be separately compared and the two points would be considered as the same if the sum of total difference is under a threshold. Even if the two end points were on the same *x-z* and *z-y* planes respectively, they would still be treated as the same points in 3D space. Thus, the visual or multi-dimensional inputs are required for ChatGPT to build the accurate 3D scene understanding for real-world operation. **Fig.20** showed the influence of the constraint caused by using different sizes of pipe. The pipeline could only get approach to the mandatory points but not pass them.

TABLE II
THE COUNTING RESULTS OF EXPERIMENTS ON PASS POINTS TASK WITH TWO CONDITIONS.

|  | Success | Fail | Total | Success/Total | Failed/Total |
|---|---|---|---|---|---|
| Constant | 14 | 6 | 20 | 0.7 | 0.3 |
| Variable | 4 | 16 | 20 | 0.2 | 0.8 |

In conclusion, the system demonstrated superior performance under constant conditions in the second task as opposed to the first one. This can be attributed to the fact that avoiding specific points offered a greater array of potential solutions compared to passing points, resulting in a higher level of stability for the ChatGPT system. Consequently, the success rate for the second task was 1, whereas it was only 0.7 for the first task.

Considering the two tasks and two conditions derived from real-world environments, it is evident that, in contrast to study cases 1 and 2, employing ChatGPT and RoboGPT systems to address real-world construction tasks introduces additional constraints that significantly impact the stability and overall performance of the system. Furthermore, it is crucial to recognize that addressing real-world tasks encompasses not only achieving the desired objectives but also optimizing resource utilization. Consequently, future research should aim to guide the ChatGPT agent towards identifying the most efficient and effective means of resolving the problem at hand.

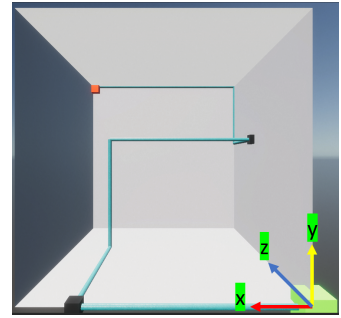

FIGURE 17. The layout of successful trial in constant condition.

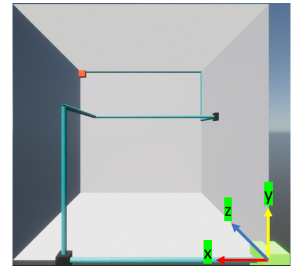

FIGURE 18. The layout of failed trial in constant condition.

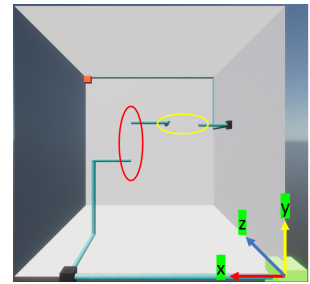

FIGURE 19. The layout of successful trial in variable condition

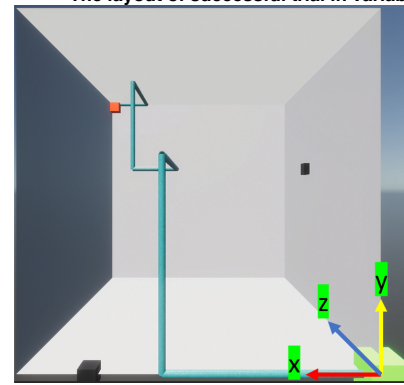

FIGURE 20. The layout of failed trial in variable condition.

## VI. CONCLUSIONS

In this paper, we presented a robotic system leveraging ChatGPT-4 for automated sequence planning in complex construction assembly tasks, such as assembling structural components of a building, installing electrical and plumbing systems, and coordinating the movement of construction equipment on site. The tasks involved a wide range of spatial constraints, including limited workspace, safe operation distances, and proper placement of components, as well as resource constraints, such as the availability of equipment and personnel. We developed a framework that allowed ChatGPT-4 to ingest relevant input data, including construction specifications, blueprints, and a list of available resources. The model was then able to generate an optimized assembly sequence plan by decomposing the tasks into logical steps, ensuring that the spatial and resource constraints were satisfied. Each step included specific instructions for the robotic system, such as the order of operations, the type and quantity of resources required, and the optimal path for movement of equipment and materials. To evaluate the effectiveness of the ChatGPT-4-based method, we compared its performance with that of two real-world construction tasks. Our results showed that the ChatGPT-4-based system has the potential to understand the background logic of a sequential task and give corresponding solution. We also used the test results from 80 trials to intuitively demonstrate the current limitation and boundary of ChatGPT agent in solving real-world tasks considering the physical constraints and resource restriction. To be prepared for assisting human workers to solve real construction problems, the ability for spatial understanding and dynamic management of ChatGPT is required to be improved.

Honestly, there are several limitations to our approach. First, we have yet to fully understand the underlying mechanisms that allow ChatGPT-4 to be used for construction task sequence planning, particularly when considering spatial and resource constraints. Second, the level of trust human workers have in the ChatGPT-4-based system remains unknown, which could impact the adoption of this technology in real-world scenarios. Lastly, ChatGPT-4's ability to process and analyze imagery data is limited, restricting its applicability in situations where visual information is crucial. Future research should focus on addressing these limitations and expanding the scope of the study. It is essential to test more construction applications to validate the robustness of the ChatGPT-4-based method and assess its performance across diverse tasks. Furthermore, investigating the reasons behind ChatGPT-4's success in construction task sequence planning will enhance our understanding of its capabilities and help improve the model. Additionally, integrating ChatGPT-4 with computer vision techniques could pave the way for a fully automated process, which would enable seamless collaboration between the language model and visual data processing systems, ultimately boosting efficiency and accuracy in construction sequence planning.

In our future work, we plan to augment our RoboGPT system with Reinforcement Learning from Human Feedback (RLHF) [59] to enhance its adaptability and robustness across a wide range of construction scenarios. To achieve this, we will design and integrate a feedback mechanism that enables the collection of human expert preferences and evaluations to guide the model's learning process. By incorporating RLHF, the RoboGPT system can iteratively update its sequence planning capabilities based on expert feedback, allowing it to better comprehend the intricacies and subtleties of construction tasks. This approach enables the system to adapt more effectively to the dynamic nature of construction projects, while also reducing reliance on large amounts of training data. Furthermore, we will develop a method for incorporating feedback from virtual simulations, which will reflect the consequences of the generated construction sequences. This additional source of feedback will enable the RoboGPT system to refine its calculations in real-time and improve its overall performance.

## ACKNOWLEDGEMENTS

This material is supported by the National Science Foundation (NSF) under grant 2128895. Any opinions, findings, conclusions, or recommendations expressed in this article are those of the authors and do not reflect the views of the NSF.

## REFERENCES

[1] A.K. Ali, O.J. Lee, H. Song, Robot-based facade spatial assembly optimization, Journal of Building Engineering, 33 (2021) 101556.
[2] A. Zhu, G. Xu, P. Pauwels, B. De Vries, M. Fang, Deep Reinforcement Learning for Prefab Assembly Planning in Robot-based Prefabricated Construction, 2021 IEEE 17th International Conference on Automation Science and Engineering (CASE), IEEE, 2021, pp. 1282-1288.
[3] S. Cai, Z. Ma, M.J. Skibniewski, S. Bao, Construction automation and robotics for high-rise buildings over the past decades: A comprehensive review, Advanced Engineering Informatics, 42 (2019) 100989.
[4] C.-J. Liang, V.R. Kamat, C.C. Menassa, Teaching robots to perform quasi-repetitive construction tasks through human demonstration, Automation in Construction, 120 (2020) 103370.
[5] Y. Shi, J. Du, D.A. Worthy, The impact of engineering information formats on learning and execution of construction operations: A virtual reality pipe maintenance experiment, Automation in Construction, 119 (2020) 103367.
[6] Z. Abuwarda, T. Hegazy, Work-package planning and schedule optimization for projects with evolving constraints, Journal of Computing in Civil Engineering, 30 (2016) 04016022.
[7] T. Kim, Y.-w. Kim, H. Cho, Dynamic production scheduling model under due date uncertainty in precast concrete construction, Journal of Cleaner Production, 257 (2020) 120527.
[8] J.P. Garcia-Sabater, J. Maheut, J.J. Garcia-Sabater, A two-stage sequential planning scheme for integrated operations planning and scheduling system using MILP: the case of an engine assembler, Flexible services and manufacturing journal, 24 (2012) 171-209.
[9] M. Suszyński, K. Peta, Assembly sequence planning using artificial neural networks for mechanical parts based on selected criteria, Applied Sciences, 11 (2021) 10414.
[10] C. Lacave, F.J. Diez, A review of explanation methods for heuristic expert systems, The Knowledge Engineering Review, 19 (2004) 133-146.
[11] A. L'heureux, K. Grolinger, H.F. Elyamany, M.A. Capretz, Machine learning with big data: Challenges and approaches, Ieee Access, 5 (2017) 7776-7797.
[12] S. Mitrović, D. Andreoletti, O. Ayoub, Chatgpt or human? detect and explain. explaining decisions of machine learning model for detecting short chatgpt-generated text, arXiv preprint arXiv:2301.13852, (2023).
[13] N. Harinarain, S. Caluza, S. Dondolo, BRICKLAYING ROBOTS IN THE SOUTH AFRICAN CONSTRUCTION INDUSTRY:

THE CONTRACTORS PERSPECTIVE, THIRTY-Seventh ANNUAL CONFERENCE, pp. 36.
[14] C.A. Cardno, Robotic rebar-tying system uses artificial intelligence, Civil Engineering Magazine Archive, 88 (2018) 38-39.
[15] S.H. Song, J.O. Choi, S. Lee, The Current State and Future Directions of Industrial Robotic Arms in Modular Construction, (2022).
[16] A.A. Apolinarska, M. Pacher, H. Li, N. Cote, R. Pastrana, F. Gramazio, M. Kohler, Robotic assembly of timber joints using reinforcement learning, Automation in Construction, 125 (2021) 103569.
[17] L. Gardner, P. Kyvelou, G. Herbert, C. Buchanan, Testing and initial verification of the world's first metal 3D printed bridge, Journal of constructional steel research, 172 (2020) 106233.
[18] A.E. Oke, J.O. Atofarati, S.F. Bello, Awareness of 3D printing for sustainable construction in an emerging economy, Construction Economics and Building, 22 (2022) 52-68.
[19] Y. Gao, J. Meng, J. Shu, Y. Liu, BIM-based task and motion planning prototype for robotic assembly of COVID-19 hospitalisation light weight structures, Automation in Construction, 140 (2022) 104370.
[20] N.E. Carey, P. Bardunias, R. Nagpal, J. Werfel, Validating a Termite-Inspired Construction Coordination Mechanism Using an Autonomous Robot, Frontiers in Robotics and AI, 8 (2021) 645728.
[21] Y. Ding, R. Dwivedi, R. Kovacevic, Process planning for 8-axis robotized laser-based direct metal deposition system: a case on building revolved part, Robotics and Computer-Integrated Manufacturing, 44 (2017) 67-76.
[22] J. Zhang, C. Shen, R. Li, A Robotic System Method and Rebar Construction with Off-the-Shelf Robots, Computing in Civil Engineering 20212022, pp. 1204-1211.
[23] T. Osa, M. Aizawa, Deep reinforcement learning with adversarial training for automated excavation using depth images, IEEE Access, 10 (2022) 4523-4535.
[24] H. Ardiny, S. Witwicki, F. Mondada, Construction automation with autonomous mobile robots: A review, 2015 3rd RSI International Conference on Robotics and Mechatronics (ICROM), IEEE, 2015, pp. 418-424.
[25] P. Adami, P.B. Rodrigues, P.J. Woods, B. Becerik-Gerber, L. Soibelman, Y. Copur-Gencturk, G. Lucas, Impact of VR-based training on human–robot interaction for remote operating construction robots, Journal of Computing in Civil Engineering, 36 (2022) 04022006.
[26] S. Alatartsev, S. Stellmacher, F. Ortmeier, Robotic task sequencing problem: A survey, Journal of intelligent & robotic systems, 80 (2015) 279-298.
[27] T. Cao, A.C. Sanderson, AND/OR net representation for robotic task sequence planning, IEEE Transactions on Systems, Man, and Cybernetics, Part C (Applications and Reviews), 28 (1998) 204-218.
[28] Y. Xing, G. Chen, X. Lai, S. Jin, J. Zhou, Assembly sequence planning of automobile body components based on liaison graph, Assembly automation, 27 (2007) 157-164.
[29] A. Yassine, An introduction to modeling and analyzing complex product development processes using the design structure matrix (DSM) method, Urbana, 51 (2004) 1-17.
[30] I.-M. Chen, G. Yang, Configuration independent kinematics for modular robots, Proceedings of IEEE International Conference on Robotics and Automation, IEEE, 1996, pp. 1440-1445.
[31] I. Craig, Blackboard systems, Intellect Books1995.
[32] C.P. Chen, Y.-H. Pao, An integration of neural network and rule-based systems for design and planning of mechanical assemblies, IEEE transactions on systems, man, and cybernetics, 23 (1993) 1359-1371.
[33] T. Vossen, M. Ball, A. Lotem, D. Nau, Applying integer programming to AI planning, The Knowledge Engineering Review, 15 (2000) 85-100.
[34] N. Sundström, O. Wigström, P. Falkman, B. Lennartson, Optimization of operation sequences using constraint programming, IFAC Proceedings Volumes, 45 (2012) 1580-1585.
[35] Z.-F. LIU, D. Hu, X. GAO, J.-D. ZHANG, Product Disassembly Sequence Planning Based on Greedy Algorithm, China Mechanical Engineering, 22 (2011) 2162.
[36] H. Shan, S. Li, D. Gong, P. Lou, Genetic simulated annealing algorithm-based assembly sequence planning, 2006 International Technology and Innovation Conference (ITIC 2006), IET, 2006, pp. 1573-1579.
[37] W. Li, S. Ong, A. Nee, Optimization of process plans using a constraint-based tabu search approach, International Journal of Production Research, 42 (2004) 1955-1985.
[38] H. Xia, X. Li, L. Gao, A hybrid genetic algorithm with variable neighborhood search for dynamic integrated process planning and scheduling, Computers & Industrial Engineering, 102 (2016) 99-112.
[39] A. Berry, P. Heggernes, G. Simonet, The minimum degree heuristic and the minimal triangulation process, Graph-Theoretic Concepts in Computer Science: 29th International Workshop, WG 2003. Elspeet, The Netherlands, June 19-21, 2003. Revised Papers 29, Springer, 2003, pp. 58-70.
[40] F. Yang, K. Gao, I.W. Simon, Y. Zhu, R. Su, Decomposition methods for manufacturing system scheduling: A survey, IEEE/CAA Journal of Automatica Sinica, 5 (2018) 389-400.
[41] H.-E. Tseng, C.-C. Chang, S.-C. Lee, Y.-M. Huang, A block-based genetic algorithm for disassembly sequence planning, Expert Systems with Applications, 96 (2018) 492-505.
[42] Z. Han, Y. Wang, D. Tian, Ant colony optimization for assembly sequence planning based on parameters optimization, Frontiers of Mechanical Engineering, 16 (2021) 393-409.
[43] L.A. Bewoor, V.C. Prakash, S.U. Sapkal, Production scheduling optimization in foundry using hybrid Particle Swarm Optimization algorithm, Procedia Manufacturing, 22 (2018) 57-64.
[44] J.F. Arinez, Q. Chang, R.X. Gao, C. Xu, J. Zhang, Artificial intelligence in advanced manufacturing: Current status and future outlook, Journal of Manufacturing Science and Engineering, 142 (2020).
[45] M.P. Strub, J.D. Gammell, Adaptively Informed Trees (AIT): Fast Asymptotically Optimal Path Planning through Adaptive Heuristics, 2020 IEEE International Conference on Robotics and Automation (ICRA), IEEE, 2020, pp. 3191-3198.
[46] M.J. Souza, I.M. Coelho, S. Ribas, H.G. Santos, L.H.d.C. Merschmann, A hybrid heuristic algorithm for the open-pit-mining operational planning problem, European Journal of Operational Research, 207 (2010) 1041-1051.
[47] H. Sun, H.V. Burton, H. Huang, Machine learning applications for building structural design and performance assessment: State-of-the-art review, Journal of Building Engineering, 33 (2021) 101816.
[48] C.F. Caiafa, Z. Sun, T. Tanaka, P. Marti-Puig, J. Solé-Casals, Machine learning methods with noisy, incomplete or small datasets, MDPI, 2021, pp. 4132.
[49] L. Floridi, M. Chiriatti, GPT-3: Its nature, scope, limits, and consequences, Minds and Machines, 30 (2020) 681-694.
[50] J. Devlin, M.-W. Chang, K. Lee, K. Toutanova, Bert: Pre-training of deep bidirectional transformers for language understanding, arXiv preprint arXiv:1810.04805, (2018).
[51] M.R. Ciosici, L. Derczynski, Training a T5 Using Lab-sized Resources, arXiv preprint arXiv:2208.12097, (2022).
[52] A. Vaswani, N. Shazeer, N. Parmar, J. Uszkoreit, L. Jones, A.N. Gomez, Ł. Kaiser, I. Polosukhin, Attention is all you need, Advances in neural information processing systems, 30 (2017).
[53] W. Huang, F. Xia, T. Xiao, H. Chan, J. Liang, P. Florence, A. Zeng, J. Tompson, I. Mordatch, Y. Chebotar, Inner monologue: Embodied reasoning through planning with language models, arXiv preprint arXiv:2207.05608, (2022).
[54] S.A. Prieto, E.T. Mengiste, B. García de Soto, Investigating the use of ChatGPT for the scheduling of construction projects, Buildings, 13 (2023) 857.
[55] I. Singh, V. Blukis, A. Mousavian, A. Goyal, D. Xu, J. Tremblay, D. Fox, J. Thomason, A. Garg, Progprompt: Generating situated robot task plans using large language models, arXiv preprint arXiv:2209.11302, (2022).
[56] S. Vemprala, R. Bonatti, A. Bucker, A. Kapoor, Chatgpt for robotics: Design principles and model abilities, (2023).

[57] Y. Bang, S. Cahyawijaya, N. Lee, W. Dai, D. Su, B. Wilie, H. Lovenia, Z. Ji, T. Yu, W. Chung, A multitask, multilingual, multimodal evaluation of chatgpt on reasoning, hallucination, and interactivity, arXiv preprint arXiv:2302.04023, (2023).
[58] S. Badini, S. Regondi, E. Frontoni, R. Pugliese, Assessing the Capabilities of ChatGPT to Improve Additive Manufacturing Troubleshooting, Advanced Industrial and Engineering Polymer Research, (2023).
[59] P.F. Christiano, J. Leike, T. Brown, M. Martic, S. Legg, D. Amodei, Deep reinforcement learning from human preferences, Advances in neural information processing systems, 30 (2017).